%% file: main.tex
\documentclass[10pt,twocolumn,letterpaper]{article}

\usepackage{cvpr}              


\definecolor{cvprblue}{rgb}{0.21,0.49,0.74}
\usepackage[pagebackref,breaklinks,colorlinks,
  linkcolor=black,
  citecolor=black,
  urlcolor=cvprblue,
  filecolor=black,
  menucolor=black,
  runcolor=black,
  anchorcolor=black
]{hyperref}
\usepackage[accsupp]{axessibility}  

\usepackage{url}
\usepackage{algorithm2e}

\usepackage[utf8]{inputenc} 
\usepackage[T1]{fontenc}    
\usepackage{url}            
\usepackage{booktabs}       
\usepackage{amsfonts}       
\usepackage{nicefrac}       
\usepackage{microtype}      

\usepackage{graphicx}
\usepackage{subcaption}

\usepackage{listings}

\usepackage{algpseudocode}
\usepackage{amsmath}
\AtEndPreamble{
    \usepackage[capitalize]{cleveref}
    \Crefname{section}{Section}{Sections}
    \Crefname{table}{Table}{Tables}
}
\usepackage{enumitem}
\usepackage{tabularx}
\usepackage{xcolor}  
\usepackage{amsmath} 

\usepackage{wrapfig}
\usepackage{multirow}

\usepackage{pifont}
\newcommand{\cmark}{\textcolor{green}{\ding{51}}} 
\newcommand{\xmark}{\textcolor{red}{\ding{55}}}   

\newcommand{\rewrite}[1]{\textcolor{black}{#1}}

\usepackage{comment}

\usepackage[bottom]{footmisc}  
\usepackage{graphicx}

\usepackage{booktabs} 
\usepackage{caption} 
\usepackage{array} 

\newcommand{\cparagraph}[1]{{\noindent\textbf{#1}\quad}}
\usepackage[normalem]{ulem}

\def\paperID{11518} 
\def\confName{CVPR}
\def\confYear{2025}

\title{Ouroboros3D: Image-to-3D Generation via 3D-aware Recursive Diffusion}

\author{
    Hao Wen\textsuperscript{1, 2}\textcolor{black}{\footnotemark[1]}
     \;
    Zehuan Huang\textsuperscript{1,3}\textcolor{black}{\footnotemark[1]} \;
    Yaohui Wang\textsuperscript{2} \;
    Xinyuan Chen\textsuperscript{2} \;
    Lu Sheng\textsuperscript{1}\textcolor{black}{\footnotemark[2]} \\
    {\textsuperscript{1}School of Software, Beihang University \quad
    \textsuperscript{2}Shanghai AI Laboratory \quad
    \textsuperscript{3}VAST \quad
    }\\
}

\begin{document}

\twocolumn[\maketitle\vspace{-1em}\input{sec/teaser}\bigbreak]

\footnotetext[1]{Equal contribution}
\footnotetext[2]{Corresponding author}

\input{sec/0_abstract}

\input{sec/1_intro}

\input{sec/2_realated_work}

\input{sec/3_method}
\input{sec/4_Exp}

\input{sec/5_con}

{
    \small
    \bibliographystyle{ieeenat_fullname}
    \bibliography{main}
}


\end{document}

%% file: sec/teaser.tex
\begin{center}
\centering
\includegraphics[width=0.95\textwidth]{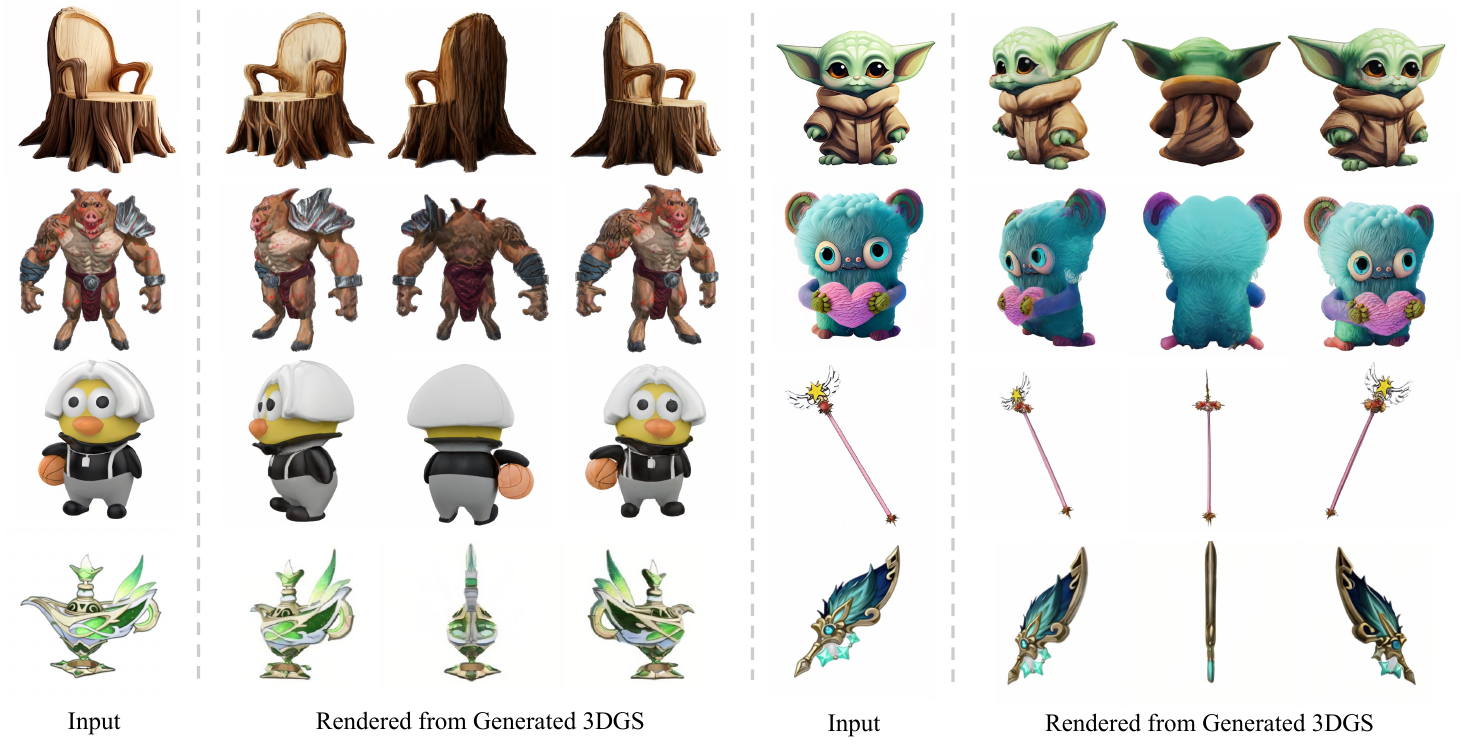}
\captionof{figure}{\textbf{Ouroboros3D} generates multi-view consistent images and high-quality 3D models from single images using 3D-aware recursive diffusion, introducing a novel \textbf{3D-aware feedback} mechanism that involves iterative cycles of multi-view denoising and reconstruction.}
\label{fig:teaser_vis}
\end{center}

%% file: sec/0_abstract.tex
\begin{abstract}
Existing image-to-3D creation methods typically split the task into two stages: multi-view image generation and 3D reconstruction, leading to two main limitations:
(1) In multi-view generation stage, the multi-view generated images present a challenge to preserving 3D consistency; (2) In the 3D reconstruction stage, a domain gap exists between the real training data and the images generated during the inference process.
To address these issues, we propose Ouroboros3D, 
an end-to-end trainable framework that integrates multi-view generation and 3D reconstruction into a recursive diffusion process through feedback mechanism. 
Our framework operates through iterative cycles where each cycle consists of a feedback denoising process and a reconstruction step. By incorporating a 3D-aware feedback mechanism, our multi-view generative model 
leverages the explicit 3D geometric information (e.g. texture, position) 
from the feedback of reconstruction results of the previous process 
as conditions, thus modeling consistency at the 3D geometric level.
Furthermore, through joint training of both the multi-view generative and reconstruction models, we alleviate reconstruction stage domain gap
and enable mutual enhancement within the recursive process. 
Experimental results demonstrate that Ouroboros3D outperforms methods that treat these stages separately and those that combine them only during inference, achieving superior multi-view consistency and producing 3D models with higher geometric realism. Please see the project page at \url{https://costwen.github.io/Ouroboros3D/}
\end{abstract}

%% file: sec/1_intro.tex
\section{Introduction}
\label{sec:intro}

\begin{figure*}[t]
    \centering
    \includegraphics[width=\linewidth]{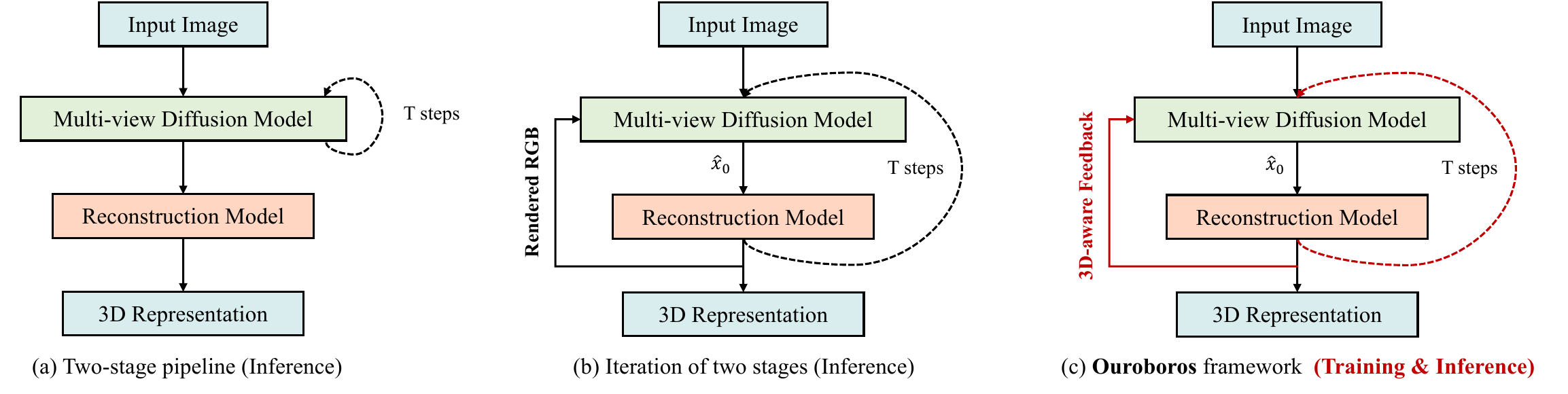}
    \caption{\textbf{Concept comparison} between Ouroboros3D and previous two-stage methods. Instead of separating multi-view diffusion model and reconstruction model, our framework involves joint training and inference of these two models, which are established into a recursive diffusion process.}
    \label{fig:concept-comparison}
\end{figure*}

Creating 3D content from a single image have achieved rapid progress in recent years with the adoption of large-scale 3D datasets~\citep{deitke2023objaverse,deitke2024objaversexl,wu2023omniobject3d} and generative models~\citep{pmlr-v37-sohl-dickstein15, ho2020denoising, song2020score}. A body of research~\citep{liu2023zero123,shi2023mvdream,liu2023syncdreamer, kwak2023vivid, huang2023epidiff,tang2024mvdiffusion++,voleti2024sv3d, long2023wonder3d} has focused on \rewrite{multi-view generation}, fine-tuning pretrained image or video diffusion models on 3D datasets to enable consistent multi-view synthesis. These methods demonstrate generalizability and produce promising results. Another group of works~\citep{hong2023lrm, tang2024lgm, xu2024grm,wang2024crm, xu2024instantmesh} propose generalizable reconstruction models, to generate 3D representation from one or few views in a feed-forward process, leading to efficient image-to-3D creation.

Since single-view reconstruction models~\citep{hong2023lrm} trained on 3D datasets~\citep{deitke2023objaverse,yu2023mvimgnet} lack generalizability and often produce blurring at unseen viewpoints, several works~\citep{li2023instant3d,tang2024lgm,wang2024crm,xu2024instantmesh} combine multi-view diffusion models and feed-forward reconstruction models, so as to extend the reconstruction stage to sparse-view input, boosting the reconstruction quality.
As shown in \cref{fig:concept-comparison}, these methods split 3D generation into two stages: multi-view synthesis and 3D reconstruction. By combining generalizable multi-view diffusion models and robust sparse-view reconstruction models, such pipelines achieve high-quality image to 3D generation. However, combining the two independently designed models introduces a significant \rewrite{limitation} to the reconstruction model. Data bias manifests primarily in two aspects: 
\rewrite{(1) In multi-view generation stage, multi-view diffusion model is optimized at the image level, not in 3D space, complicating the assurance of geometric consistency. (2) In 3D reconstruction stage, reconstruction model is trained primarily on synthetic data with limited real data, suffering from domain gap when processing generated multi-view images.}


Recent works have explored several mechanisms to enhance multi-view consistency. Carve3D~\citep{xie2024carve3d} employs a RL-based fine-tuning algorithm~\citep{black2023training}, applying a multi-view consistency metric to enhance the multi-view image generation.
However, the challenge of data limitation has not been well-addressed, leading to poor reconstruction quality.
On the other hand, IM-3D~\citep{im3d2024} and VideoMV~\citep{zuo2024videomv} aggregate the rendered views of the reconstructed 3D model into multi-view synthesis during inference by adopting re-sampling strategy in the denoising loop.
However, on the overall image-to-3D pipeline, it (a) lacks joint training and (b) inability to use space information hinder its capacity to fully leverage 3D-aware knowledge and unify the two stages.
Moreover, these methods fail to address the \rewrite{domain gap} between generated multi-view images and training data in 3D reconstruction stage, and the use of biased information from few-shot reconstructed 3D models can result in multi-view outputs misaligned with the input image (see \cref{fig:quali-multi-view}).


In this paper, we introduce Ouroboros3D, a novel image-to-3D framework that seamlessly integrates multi-view generation with 3D reconstruction within a recursive diffusion process, as depicted in \cref{fig:teaser}. 
To facilitate the modeling of multi-view consistency, we propose a 3D-aware feedback mechanism, where our multi-view diffusion model utilizes 3D-aware maps rendered by the reconstruction module from the previous timestep as additional conditions during the denoising phase. Leveraging 3D-aware feedback from reconstructed representations, our model produces images with enhanced geometric consistency.
To address the  \rewrite{domain gap between the reconstruction stage training data and generated multi-view data}, we involve joint training of the multi-view diffusion model and reconstruction model.
During training, the reconstruction model utilizes images restored by the diffusion process rather than original images.
\rewrite{Joint training two module} not only reduces the \rewrite{domain gap in}  the reconstruction stage, increasing the diversity of the reconstruction, but also enhances the capability of diffusion model to generate images better suited for few-shot reconstruction \rewrite{(consider reconstruction model as a criterion to supervise multi-view diffusion model)}, making the two stages mutually beneficial in the multi-step iterative diffusion process.
The 3D-aware recursived diffusion, with the integration of the two stages, facilitates adaptive refinement of outputs through mutual feedback, enhancing inference stability and reducing data bias.


Experimental results on the GSO dataset~\citep{downs2022googleso} show that our framework outperforms separation of these stages and existing methods~\citep{zuo2024videomv} that combine the stages at the inference phase. 

Our key contributions are as follows:
\begin{itemize}[left=0pt,itemsep=0pt]
\item We introduce a image-to-3D creation framework Ouroboros3D, which integrates multi-view generation and 3D reconstruction into a recursive diffusion process. The framework is highly extensible and can accommodate various multi-view generation networks and reconstruction networks.
\item Ouroboros3D employs 
3D-aware feedback mechanism, using rendered maps to guide the multi-view generation, ensuring better geometric consistency and robustness.
\item We  conducte extensive experiments to demonstrate that Ouroboros3D significantly reduces \rewrite{reconstruction inference domain gap} and outperforms both the method that separates the two stages and the method that combines them only at inference time. 
\end{itemize}

%% file: sec/2_realated_work.tex
\section{Related Work}
\label{sec:related_work}

\paragraph{Image/Video Diffusion for Multi-view Generation}
Diffusion models~\citep{
rombach2022high, saharia2022imagen, podell2023sdxl, sauer2024sdv3,2022videodiffusionmodel,ho2022imagenvideo,singer2022makeavideo,svd2023,wang2023lavie, hu2024animate,ma2024latte,sora2024,dong2024tela,huang2024parts2whole} have demonstrated their powerful generative capabilities in image and video generation fields. Current research~\citep{liu2023zero123,shi2023mvdream,liu2023syncdreamer, kwak2023vivid, huang2023epidiff,tang2024mvdiffusion++,voleti2024sv3d, long2023wonder3d,zheng2023free3d,huang2024mvadapter,zhang2021writeananimation} fine-tunes pretrained image/video diffusion models on 3D datasets like Objaverse~\citep{deitke2023objaverse} and MVImageNet~\citep{yu2023mvimgnet}. Zero123~\citep{liu2023zero123} introduces relative view condition to image diffusion models, enabling novel view synthesis from a single image and preserving generalizability. Based on it, methods like SyncDreamer~\citep{liu2023syncdreamer}, ConsistNet~\citep{yang2023consistnet} and EpiDiff~\citep{huang2023epidiff} design attention modules to generate consistent multi-view images. These methods fine-tuned from image diffusion models produce generally promising results. 
By considering multi-view images as consecutive frames of a video (e.g., orbiting camera views), it naturally leads to the idea of applying video generation models to 3D generation~\citep{voleti2024sv3d}. However, since the diffusion model is not explicitly modeled in 3D space, the generated multi-view images often struggle to achieve consistent and robust details.

\paragraph{Image to 3D Reconstruction}
Recently, the task of reconstructing 3D objects has evolved from traditional multi-view reconstruction methods~\citep{mildenhall2021nerf, barron2021mip, muller2022instant,3dgs2023} to feed-forward reconstruction models~\citep{hong2023lrm,jiang2023leap, zou2023triplane, tang2024lgm, xu2024grm,wang2024crm, xu2024instantmesh,huang2024midi}. Ultilizing one or few shot as input, these highly generalizable reconstruction models synthesize 3D representation, enabling the rapid generation of 3D objects. LRM~\citep{hong2023lrm} proposes a transformer-based model to effectively map image tokens to 3D triplanes. Instant3D~\citep{li2023instant3d} further extends LRM to sparse-view input, significantly boosting the reconstruction quality. LGM~\citep{tang2024lgm} and GRM~\citep{xu2024grm} replace the triplane representation with 3D Gaussians~\citep{3dgs2023} to enjoy its superior rendering efficiency. CRM~\citep{wang2024crm} and InstantMesh~\citep{xu2024instantmesh} optimize on the mesh representation for high-quality geometry and texture modeling. These reconstrucion models built upon convolutional network architecture or transformer backbone, have led to efficient image-to-3D creation.
\begin{figure}
    \centering
    \includegraphics[width=\linewidth]{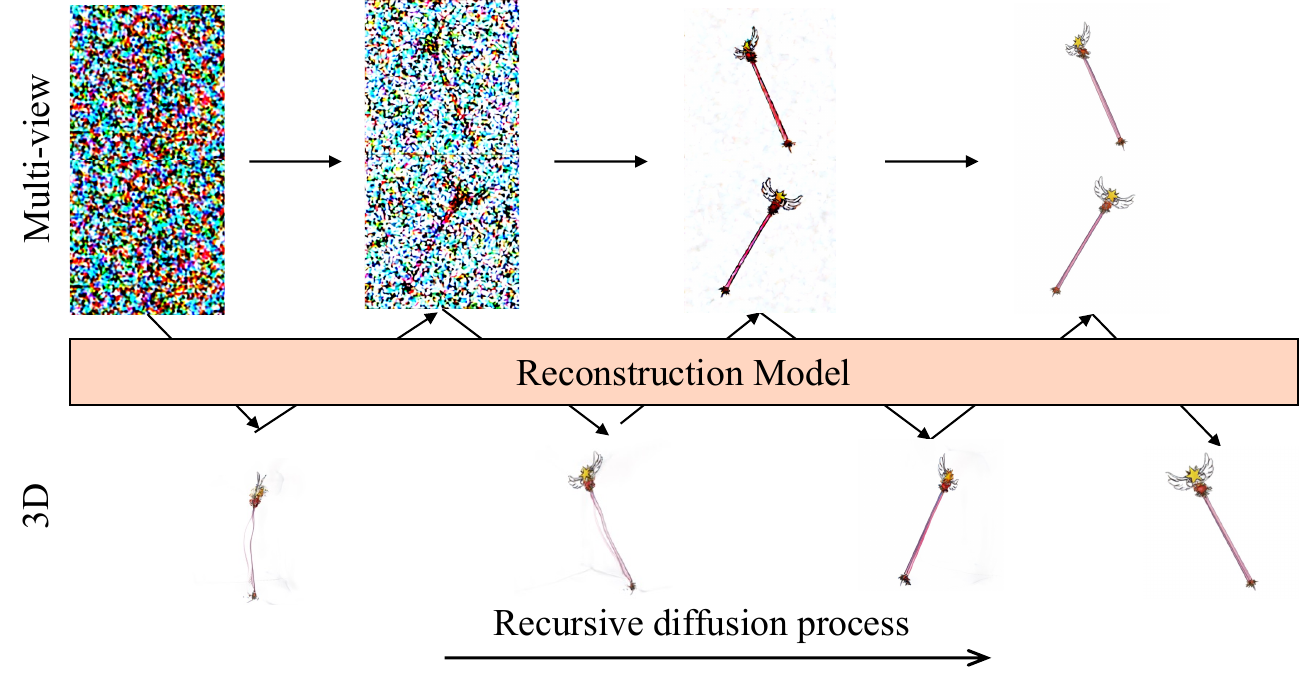}
    \caption{\textbf{Overview of 3D-aware recursive diffusion.} During multi-view denoising, the diffusion model uses 3D-aware maps rendered by the reconstruction module at the previous step as conditions.}
    \label{fig:teaser}
\end{figure}
\paragraph{Pipelines of 3D Generation}
Early works propose to distill knowledge of image prior to create 3D models via Score Distillation Sampling (SDS)~\citep{poole2022dreamfusion,lin2023magic3d,threestudio2023}, limited by the low speed of per-scene optimization.  DMV3D~\citep{xu2023dmv3d} employs a 3D reconstruction model as the 2D multi-view denoiser in a multiview diffusion framework, to achieve generic end-to-end 3D generation. However, it fails to utilize the advanced features of pre-existing image or video diffusion models, and training from scratch on 3D data limits its generalization. Several works~\citep{liu2023syncdreamer,huang2023epidiff,long2023wonder3d,im3d2024} fine-tune image diffusion models to generate multi-view images, which are then utilized for 3D shape and appearance recovery with traditional reconstruction methods~\citep{wang2021neus,3dgs2023}. More recently, several works~\citep{li2023instant3d,tang2024lgm,wang2024crm,xu2024instantmesh,zuo2024videomv} involve both multi-view diffusion models and feed-forward reconstruction models in the generation process. Such pipelines attempt to combine the processes into a cohesive two-stage approach, thus achieving highly generalizable and high-quality single-image to 3D generation. The multi-view diffusion model, lacking explicit 3D modeling, struggles to ensure strong consistency, resulting in data deviations between the testing and training phases. In contrast, we propose a unified pipeline that integrates these stages through a self-conditioning mechanism during training, enhanced by 3D-aware feedback to achieve high consistency.

%% file: sec/3_method.tex
\section{Method}

\begin{figure*}
    \centering
    \includegraphics[width=0.9\linewidth]{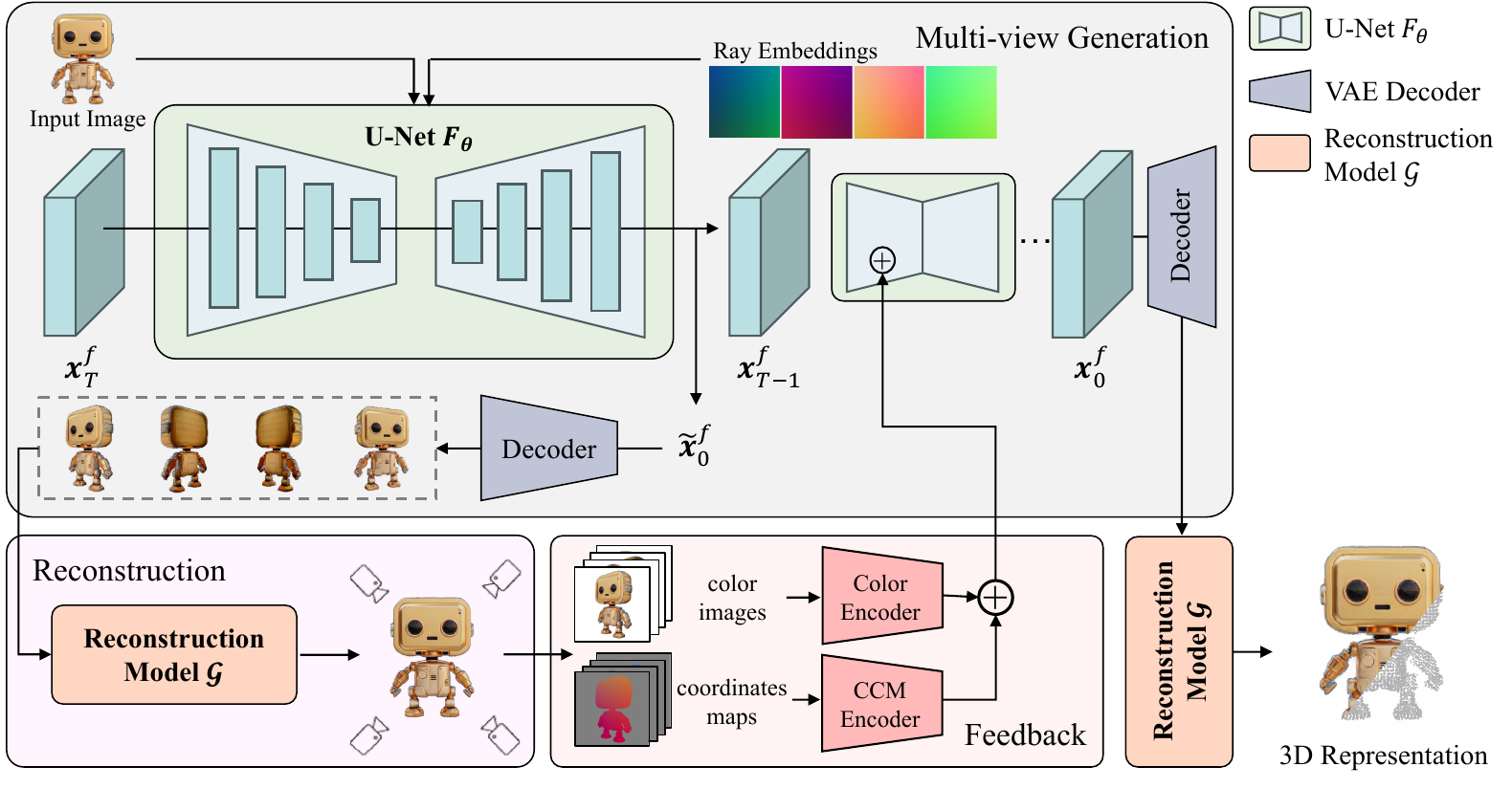}
    \caption{\textbf{Overview of Ouroboros3D.} We adopt a video diffusion model as the multi-view generator by incorporating the input image and relative camera poses. In the denoising sampling loop, we decode the predicted $\mathbf{\widetilde{x}}_{0}^{f}$ to noise-corrupted images, which are then used to recover 3D representation by a feed-forward reconstruction model. Then the rendered color images and coordinates maps are encoded and fed into the next denoising step. At inference, the 3D-aware denoising sampling strategy iteratively refines the images by incorporating feedback from the reconstructed 3D into the denoising loop, enhancing multi-view consistency and image quality.}
    \label{framework}
\end{figure*}

Given a single image, Ouroboros3D aims to generate multiview-consistent images with a reconstructed 3D Gaussion model. To reduce the \rewrite{domain gap} and improve robustness of the generation, our framework integrates multi-view synthesis and 3D reconstruction in a recursive diffusion process. As illustrated in \cref{framework}, the proposed framework involves a video diffusion model (SVD~\citep{svd2023}) as multi-view generator (refer to \cref{videodiffusion}) and a feed-forward reconstruction model to recover a 3D Gaussian Splatting (refer to \cref{feed-forward model}. Moreover, we introduce a self-conditioning mechanism, feeding the 3D-aware information obtained from the reconstruction module back to the multi-view generation process (refer to \cref{3dfeedback}). The 3D-aware recursive diffusion strategy iteratively refines the multi-view images and the 3d model, enhancing the final production.

\subsection{Video Diffusion Model as Multiview Generator}
\label{videodiffusion}
Recent video diffusion models~\citep{voleti2024sv3d,sora2024,gao2024cat3d} have demonstrated a remarkable capability to generate 3D-aware videos. We employs the well-known Stable Video Diffusion (SVD) Model as our multi-view generator, which generates videos from an image input. In our framework, we set the number of the generated frames $f$ to 8.

We enhance the video diffusion model with camera control $c$ to generate images from different viewpoints. Traditional methods encode camera positions at the frame level, which results in all pixels within one view sharing the same positional encoding~\citep{liu2023zero,voleti2024sv3d}. Building on the innovations of previous work~\citep{huang2023epidiff,zheng2023free3d}, we integrate the camera condition $c$ into the denoising network by parameterizing the rays $\mathbf{r} = (o, o \times d)$. Specifically, we use two-layered MLP to inject Plücker ray embeddings for each latent pixel, enabling precise positional encoding at the pixel level. This approach allows for more detailed and accurate 3D rendering, as pixel-specific embedding enhances the model's ability to handle complex variations in depth and perspective across the video frames.

Our multi-view diffusion model differs from existing two-stage methods in that it does not independently complete all denoising steps. Instead, within the denoising sampling loop, we obtain the predicted $\widetilde{\mathbf{x}}_{0}^{f}$ at each timestep, where 
$f$ indicates the frame number, which is then utilized for subsequent 3D reconstruction. The rendered maps are employed as conditions to guide the next denoising step. At each sampling step,we reparameterize the output from the denoising network $F_\theta$ to transform it into $\widetilde{\mathbf{x}}_{0}^{f}$. we apply the following formula to process the noising images \( c_{\text{in}}(\sigma)\mathbf{x}^f \) and the associated noise level \( c_{\text{noise}}(\sigma) \):
\begin{equation}
\label{reparam}
       \Tilde{\mathbf{x}}_0^f = c_{\text{skip}}(\sigma)\mathbf{x}^f + c_{\text{out}}(\sigma)F_\theta(c_{\text{in}}(\sigma)\mathbf{x}^f; c_{\text{noise}}(\sigma)).
\end{equation}
where \( \sigma \) indicates the standard deviation of the noise, $c_{\text{skip}}$ is a parameter that controls how much of the original $\mathbf{x}_0^f$ is retained. This operation adjusts the output of \( F_\theta \) to $\widetilde{\mathbf{x}}_{0}^{f}$, which will be decoded into images and passed to the subsequent 3D reconstruction module.

\subsection{Feed-Forward Reconstruction Model}
\label{feed-forward model}

In the Ouroboros3D framework, the feed-forward reconstruction model is designed to recover 3D models from pre-generated multi-view images, which can be images decoded from straightly predicted $\widetilde{\mathbf{x}}_{0}^{f}$, or completely denoised images. We utilize Large Multi-View Gaussian Model (LGM)~\citep{tang2024lgm} $\mathcal{G}$ as our reconstruction module due to its real-time rendering capabilities that benefit from 3D representation of Gaussian Splatting. 

We pass four specific views from the reparameterized output $\widetilde{\mathbf{x}}_{0}^{f}$ to the Large Gaussian Model (LGM) for 3D Gaussian Splatting reconstruction. To enhance the performance of LGM, particularly its sensitivity to different noise levels $c_\text{noise}(\sigma)$ and image details, we introduce a zero-initialized time embedding layer within the original U-Net structure of the LGM.  During training, we use 8 multi-view images encircling the object and 4 random views to supervise the model.

The loss function employed for the fine-tuning of the LGM is articulated as follows:
\begin{equation}
    \begin{aligned}
        \mathcal{L}_{\mathcal{G}} = {} & \mathcal{L}_{\text{rgb}}(\mathbf{I}, \mathcal{G}(\Tilde{\mathbf{x}}_0,c_\text{noise}(\sigma), \mathbf{C})) \\
        & +\lambda \mathcal{L}_{\text{LPIPS}}(\mathbf{I}, \mathcal{G}(\Tilde{\mathbf{x}}_0, c_\text{noise}(\sigma), \mathbf{C}))
\end{aligned}
\end{equation}
where $\mathbf{I}$ represents the set of supervised multiview images, $\mathbf{C}$ is the corresponding camera parameters.

Additionally, to maintain the model's reconstruction capability for normal images, we also input the model without adding noise and calculate the corresponding loss. In this case, we set $ c_\text{noise}(\sigma)$ to 0.

\subsection{3D-Aware Feedback Mechanism}
\label{3dfeedback}

We use a 3D-aware feedback mechanism, as shown in \cref{framework}, involving rendered color images and geometric maps in a denoising loop to enhance multiview consistency and facilitate cyclic adaptation. Unlike integrating multi-view generation and 3D reconstruction at the inference stage using re-sampling strategy~\citep{im3d2024,zuo2024videomv}, we jointly train these two modules to support more informative feedback. 

In practice, we obtain color images and canonical coordinates maps (CCM)~\citep{li2023sweetdreamer} from the reconstructed 3D model, and utilize them as condition to guide the next denoising step of multi-view generation.
We choose CCM over depth or normal maps because CCMs capture global vertex coordinates normalized across the entire 3D model, unlike depth maps that normalize relative to the self-view. This operation enables the rendered maps to be characterized as cross-view alignment, providing the strong guidance of more explicit cross-view geometry relationship. 

To encode color images and coordinates maps into the denoising network of multi-view generation module, we design two simple and lightweight encoders for color images and coordinates maps using a series of convolutional neural networks, like T2I-Adapter~\citep{mou2024t2iadapter}. The encoders are composed of four feature extraction blocks and three downsample blocks to change the feature resolution, so that the dimension of the encoded features is the same as the intermediate feature in the encoder of U-Net denoiser. The extracted features from the two conditional modalities are then added to the U-Net encoder at each scale.

We then introduce the proposed 3D-aware self-conditioning~\citep{2022selfcondition} strategy for both training and inference. The original multi-view denoising network $F_\theta(\mathbf{x};\sigma)$ is augmented with 3D-aware feedback, formulated as $F_\theta(\mathbf{x};\sigma,\mathcal{G}(\Tilde{\mathbf{x}}_0))$, where $\mathcal{G}(\Tilde{\mathbf{x}}_0)$ is the rendered maps of the reconstruction module.

\cparagraph{Training Strategy} The training of the 3D-aware multi-view generation network involves a probabilistic self-conditioning mechanism. During each training iteration, the network uses the rendered results from a feed-forward model as self-conditioning input with a probability of 0.5.
This approach ensures balanced learning and prevents the model from over-relying on the 3D information. We use the reconstruction module to lift multi-view information into explicit 3D representation, and render it to geometrically aligned RGB and CCM views, which subsequently guide the multi-view generation.The detail of training algorithm can be seen in supplementary material.

\cparagraph{Inference Strategy} In inference stage, the initial condition $\mathcal{G}(\Tilde{\mathbf{x}}_0)$ is set to zero. At each subsequent timestep, this 3D condition is updated based on the previous reconstruction result, then as a self feedback mechanism to guidance next denoising process. This iterative updating process refines the 3D representation, enhancing the consistency of multi-view images and improving the quality of the reconstructed 3D models. The detail of inference algorithm can be seen in supplementary material.


%% file: sec/4_Exp.tex
\section{Experiments}

\begin{table}\footnotesize
    \centering
    \caption{Quantitative comparison on the quality of generated multi-view (MV.) images and 3D representation for image-to-multiview and image-to-3D tasks.}
    \label{tab:comp-metrics}
    \begin{tabular}{cccccc}
    \toprule
     & Method &Res& PSNR$\uparrow$ & SSIM$\uparrow$ & LPIPS$\downarrow$ \\
    \midrule
    \multirow{4}{*}{MV.}
    & VideoMV~\citep{zuo2024videomv} & $256$ &  18.605 & 0.8410 & 0.1548 \\
    & SyncDreamer~\citep{liu2023syncdreamer} &$256$& 20.056 & 0.8163 & 0.1596 \\
    & SV3D~\citep{voleti2024sv3d} & $576$ & 21.042 & 0.8497 & 0.1296 \\
    & Ouroboros3D & $512$ & \textbf{21.770} & \textbf{0.8866} & \textbf{0.1093} \\
    \midrule
    \multirow{5}{*}{3D}
    & LGM~\citep{tang2024lgm} & $512$ &  17.716&0.8319& 0.1894\\
    & TripoSR~\citep{tochilkin2024triposr} &$256$& 18.481 & 0.8506 & 0.1357 \\
    & VideoMV(GS)~\citep{zuo2024videomv} &$256$ & 18.764 & 0.8449 & 0.1569 \\
    & InstantMesh ~\citep{xu2024instantmesh} &$512$ & 19.948 & 0.8727 & 0.1205 \\
    & Ouroboros3D & $512$ & \textbf{21.761} & \textbf{0.8894} & \textbf{0.1091} \\
    \bottomrule
    \end{tabular}
\end{table}

\begin{figure*}
    \centering
    \includegraphics[width=0.9\textwidth]{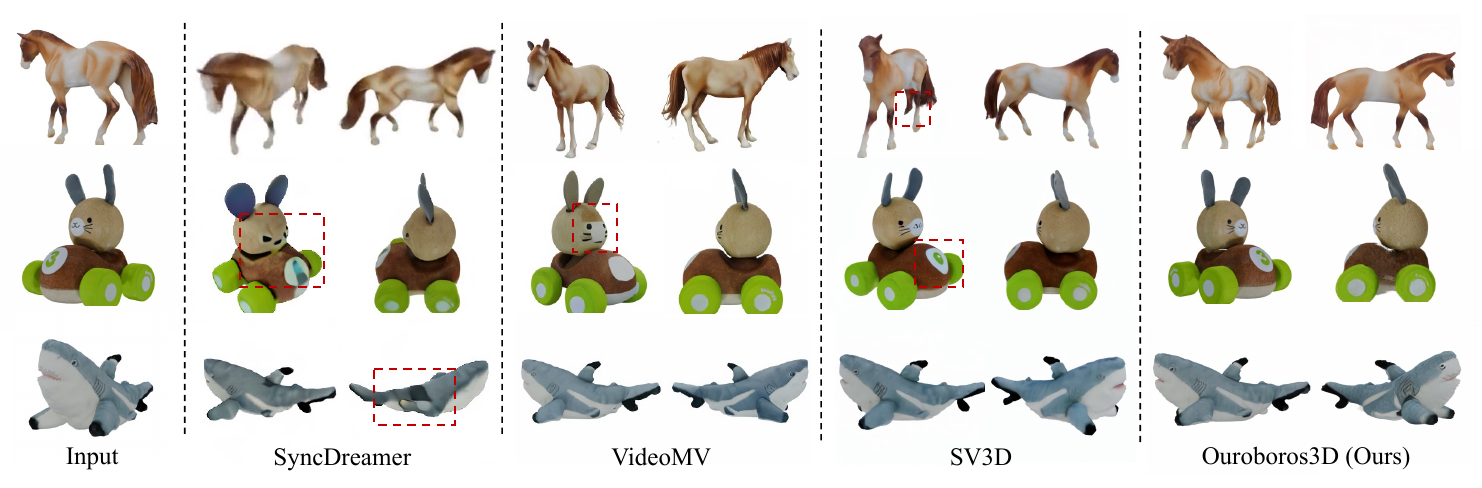}
    \caption{Qualitative comparisons of generated multi-view images. Our method achieves better consistency and quality.}
    \label{fig:quali-multi-view}
\end{figure*}
\begin{figure*}
    \centering
    \includegraphics[width=0.9\textwidth]{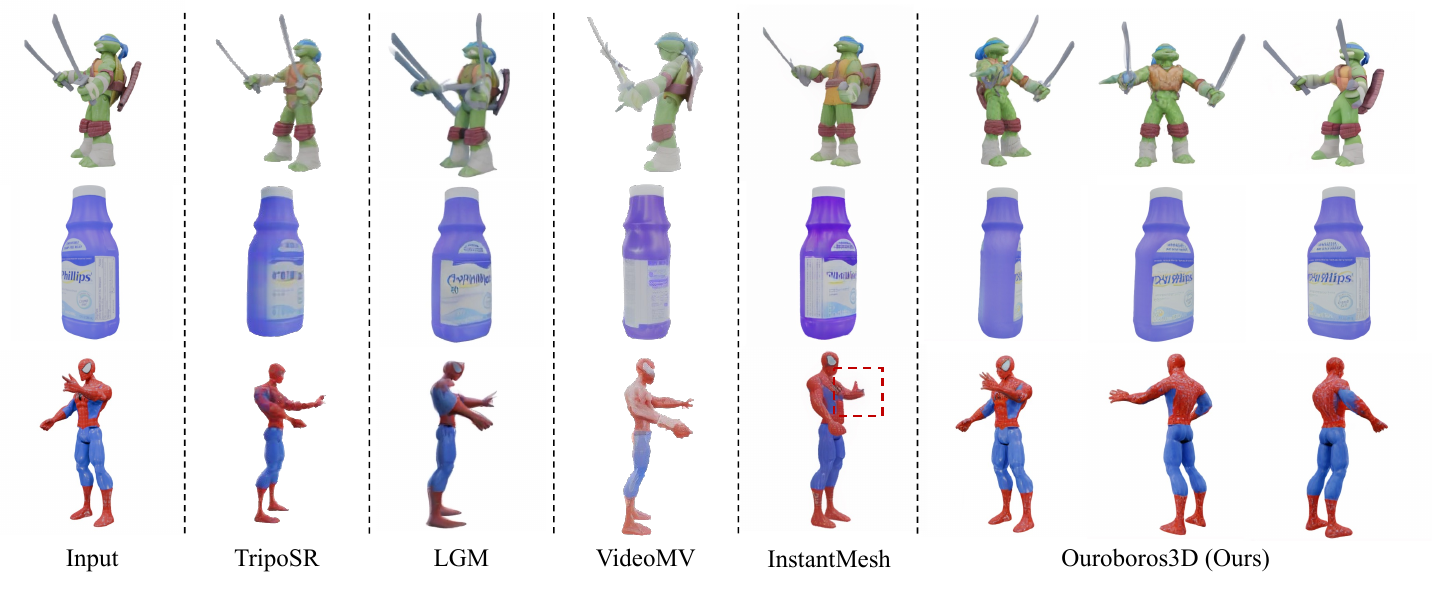}
    \caption{Qualitative comparisons for image-to-3D generation.}
    \label{fig:quali-3d}
\end{figure*}

\subsection{Implementation Details}
\label{exp-setting}
\cparagraph{Datasets}
We use a filtered subset of the Objaverse~\citep{deitke2023objaverse} dataset to train our model. Following LGM~\citep{tang2024lgm}, we implemented a rigorous filtering process to remove bad models with bad captions or missing texture. It leads to a final set of around 80K 3D objects. We render 2 16-frame RGBA orbits at $512\times 512$. For each orbit, the cameras are positioned at a randomly sampled elevation between [-5, 30] degrees. During training, we subsample any 8-frame orbit by picking any frame in one orbit as the first frame (the conditioning image), and then choose every 2nd frame after that.

We evaluate the synthesized multi-view images and reconstructed 3D Gaussian Splatting (3DGS) on the unseen GSO~\citep{downs2022googleso} dataset. We filter 100 objects to reduce redundancy and maintain diversity then render ground truth orbit videos and pick the first frame as the conditioning image.

\begin{table*}\small
    \centering
    \caption{Ablation study of different feedback mechanisms. Results show that our 3D-aware feedback mechanism lead to superior generalization performance. For implementations involving feedback, we employ the joint training strategy for model optimization.}
    \label{tab:ablation}
    \begin{tabular}{cccccccccc}
    \toprule
    Joint Training & CCM Feedback & RGB Feedback & PSNR$\uparrow$ & SSIM$\uparrow$ & LPIPS$\downarrow$ & $\Delta$PSNR$\downarrow$ & $\Delta$SSIM$\downarrow$ & $\Delta$LPIPS$\downarrow$ \\
    \midrule
    \xmark&\xmark & \xmark & 20.012 & 0.8465 & 0.1287 & 1.067 & 0.0125 & 0.0189 \\
    \cmark&\xmark & \xmark & 20.549 & 0.8651 & 0.1183 & 0.511 & 0.0094 & 0.0070 \\
    \cmark&\cmark & \xmark & 21.325 & 0.8937 & 0.1092 & 0.304 & 0.0036 & 0.0018 \\
    \cmark&\xmark & \cmark & 21.542 & 0.8871 & 0.1103 & 0.100 & 0.0101 & 0.0036 \\
    \cmark&\cmark & \cmark & \textbf{21.761} & \textbf{0.9094} & \textbf{0.0991} & \textbf{0.009} & \textbf{0.0028} & \textbf{0.0002} \\
    \bottomrule
    \end{tabular}
\end{table*}

\begin{figure*}
    \centering
    \includegraphics[width=0.9\textwidth]{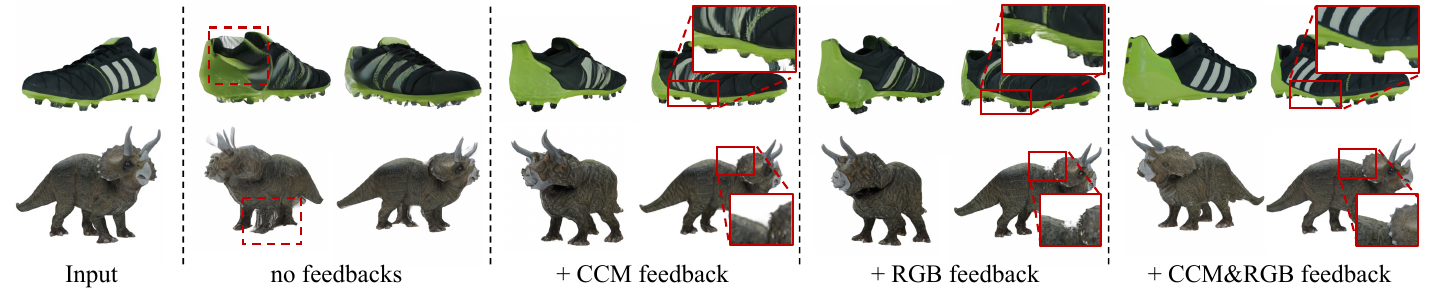}
    \caption{Qualitative ablation study on the reconstruction results with two types of feedback.}
    \label{fig:ablation}
\end{figure*}

\cparagraph{Experimental Settings}
Our Ouroboros3D is trained for 30,000 iterations using 8 A100 GPUs with a total batch size of 32. We clip the gradient with a maximum norm of 1.0. We use the AdamW optimizer with a learning rate of $1 \times 10^{-5}$ and employ FP16 mixed precision with DeepSeed\citep{rasley2020deepspeed} with Zero-2 for efficient training. At the inference stage, we set the number of sampling steps as 25, which takes about 20 seconds to generate a 3d model.

\cparagraph{Metrics}
We compare generated multi-view images and rendered views from reconstructed 3DGS with the ground truth frames, in terms of Learned Perceptual Similarity (LPIPS~\citep{lpips}), Peak Signal-to-Noise Ratio (PSNR), and Structural SIMilarity (SSIM).

\cparagraph{Baselines}
In terms of multi-view generation, we compare Ouroboros3D with SyncDreamer~\citep{liu2023syncdreamer}, SV3D~\citep{voleti2024sv3d}, VideoMV~\citep{zuo2024videomv}. For image-to-3D creation, we adopt feed-forward reconstruction models or pipelines as baseline methods, including TripoSR~\citep{tochilkin2024triposr}, LGM~\citep{tang2024lgm} and InstantMesh(Nerf)~\citep{xu2024instantmesh}, where LGM and InstantMesh adopt two-stage methods to achieve image-to-3D creation.

\subsection{Comparison with Existing Alternatives}

\cparagraph{Image-to-Multiview generation}
We compare our method with SyncDreamer~\citep{liu2023syncdreamer}, SV3D~\citep{voleti2024sv3d} and VideoMV~\citep{zuo2024videomv}, as shown in \cref{fig:quali-multi-view}. SyncDreamer and SV3D fine-tune image or video diffusion models on 3D datasets but lack explicit 3D information, often resulting in blurry textures or inconsistent details. VideoMV aggregates rendered views from reconstructed 3D models at the inference stage but fails to take into account the domain gap between two stages. Although VideoMV improves the multi-view consistency, it introduces biased information from the reconstruction stage, leading to results that are unaligned with the input image. Our Ouroboros3D uses joint training of the two stages and uses geometry and appearance feedback for multi-view generation, generating consistent and high-quality multi-view images.

\cparagraph{Image-to-3D generation} We compare our method with TripoSR~\citep{tochilkin2024triposr}, VideoMV~\citep{zuo2024videomv}, LGM~\citep{tang2024lgm} and InstantMesh~\citep{xu2024instantmesh}, as visualized in \cref{fig:quali-3d}. TripoSR struggles with high-quality geometry and appearance due to lacking large pre-trained generative models. VideoMV reconstructs 3DGS from its generated multi-view images, but its inherent biases in multiview generation can lead to misaligned textures and distorted geometries. Two-stage methods such as LGM and InstantMesh, which comprise an off-the-shelf image-to-multiview generation method followed by reconstruction models for the image-to-3D generation process, often yield incomplete geometry due to the disparity between multiview generation and 3D reconstruction. In contrast, our framework integrates multiview generation and 3D reconstruction, enhancing each module's strengths to produce high-quality 3D assets.

\begin{figure*}
    \centering
    \includegraphics[width=0.9\linewidth]{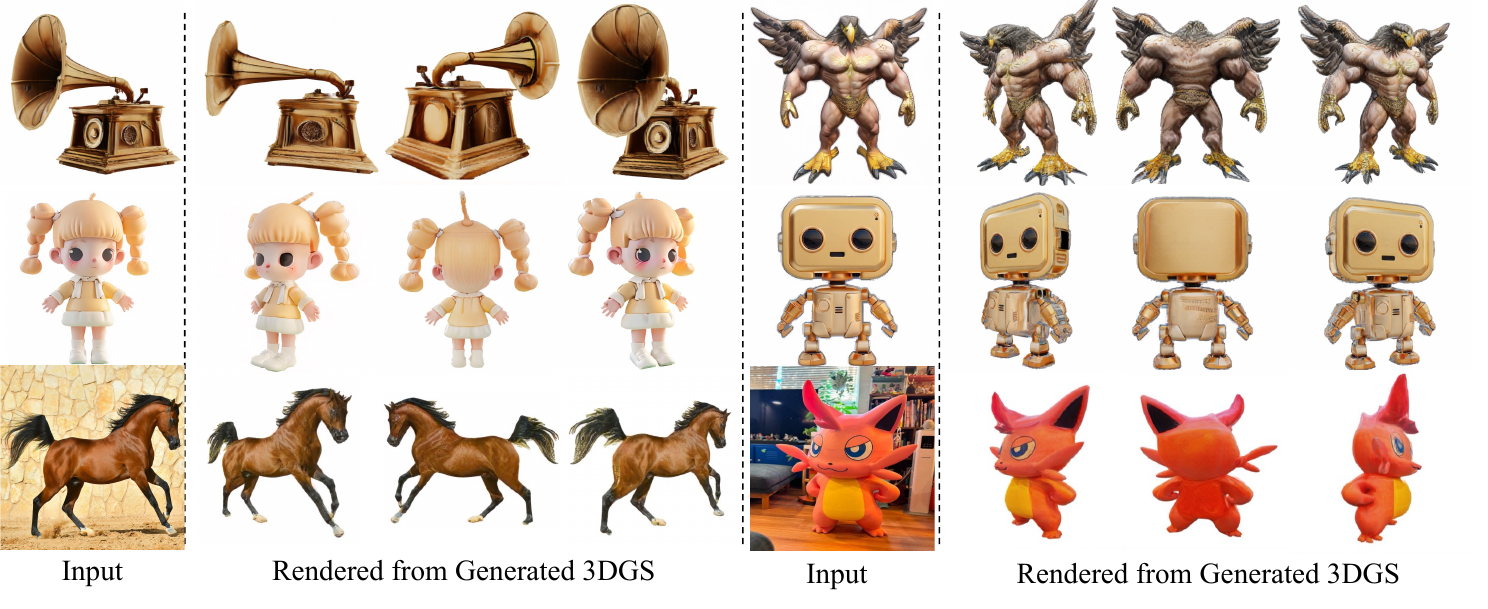}
    \caption{Ouroboros3D can generate high-quality 3D models given image inputs outside the distribution, including real world images.}
    \label{fig:ood}
\end{figure*}

\cparagraph{Generalizability} Ouroboros3D exhibits remarkable generalizability, adept at producing high-quality 3D models from images that fall outside its training distribution, including real-world images. This capability is demonstrated in the results shown in \cref{fig:ood} and supplementary materials.





\subsection{Ablation Study}

To assess the effectiveness of our 3D-aware feedback mechanism, we conducted ablation experiments on the generated 3DGS for different configurations (\cref{fig:ablation} and \cref{tab:ablation}).
We start with a base framework that does not jointly trains the multi-view generation module and the reconstruction module, or use feedback mechanism.
We then incrementally add components of our proposed approach.

\begin{table}[h]
    \centering
    \small
    \caption{Ablation on feedback steps}
    \begin{tabular}{lccc}
    \toprule
    Feedback Steps & PSNR$\uparrow$ & SSIM$\uparrow$ & LPIPS$\downarrow$ \\
    \midrule
    $0\sim 25$  & \textbf{21.761} & 0.9094 & \textbf{0.0991} \\
    $10\sim 25$ & 21.633 & \textbf{0.9099} & 0.1021 \\
    $20\sim 25$ & 20.805 & 0.8957 & 0.1101 \\
    \bottomrule
    \end{tabular}
    \label{tab:ablation_feedback}
    \small
\end{table}

The reconstructed results shown in \cref{fig:ablation} demonstrate that, only the coordinates map feedback produces blurry textures, and only the color map has poor geometric quality in fine details. Our full setting leads to superior performance in both geometry and texture.
\cref{tab:ablation} reports the quantitative results, which demonstrate significant improvements by enhancing both geometric consistency and texture details.
We also report the absolute distances of performance metrics between the generated multiviews and 3DGS. It can be observed that our framework reduces the performance difference between the generated multi-view images and 3D representation, and improves the combined performance.

We conduct the experiments on the feedback steps.
As shown in \cref{tab:ablation_feedback}, excluding feedback for the first 10 steps yields comparable results to the full setting, whereas excluding it for 20 steps reduces performance.

\begin{table}
\footnotesize
\centering
\caption{Comparison of training and inference speed. Left: training time for 1,000 steps; Right: inference time per sample.}
\label{tab:comparison}
\begin{tabular}{lc|lc}
\toprule
Setting & Training Time & Setting & Inference Time \\
\midrule
SVD & 15 min & LGM & 1.225s \\
LGM & 10 min & SV3D + LGM & 24.18s \\
Ouroboros3D & 36 min & Ouroboros3D & 25.19s \\
\bottomrule
\end{tabular}
\end{table}

\subsection{Discussion}
\label{limition}

\cparagraph{Recursive Generation Process}
We visualize the reconstruction process at different denoising steps in the supplementary materials.
Early stages show floating artifacts and distorted geometries due to multi-view inconsistency.
As denoising progresses, our recursive diffusion method gradually refines both the geometric accuracy and material properties of the reconstruction. Compare to no feedback results, the model achieves better shape and texture refinement during early stages with the feedback mechanism.

\cparagraph{Alternative 3D Representations}  Our model currently utilize 3D Gaussian splatting as the generated 3D representation, which is not as widely used as meshes. Replacing the reconstruction module with CRM~\citep{wang2024crm} or InstantMesh~\citep{xu2024instantmesh} can enable our framework to generate meshes from a single image. In addition, experiments on 3D scene dataset will also be an extension of our framework.

\subsection{Limitation}

\paragraph{Training and Inference Efficiency}
While our joint training method enhances model performance, it increases requires additional time and more GPU memory . We measure the time required for 1,000 training steps on an A100 GPU. As shown in \cref{tab:comparison}, Ouroboros3D takes longer to train than the individual components when trained separately, primarily due to the extra computations and the need for 3d feedback.

We also evaluate the inference speed of baseline methods under identical settings to ensure fairness. Despite the higher training cost, the results showed that the inference efficiency of our method is comparable to that of the baselines.
The LGM baseline employs ImageDream~\citep{wang2023imagedream} to generate $4$ views at $256 \times 256$ resolution, which are then reconstructed into a 3D Gaussian Splatting (3DGS) representation.
In contrast, our Ouroboros3D approach utilizes SVD to generate $8$ views at $512 \times 512$ resolution.
For a fair comparison, we report the inference time of "SV3D + LGM", where SV3D~\citep{voleti2024sv3d} is a multi-view generator fine-tuned from SVD.
Compared to it, the additional overhead in our method mainly stems from the feedback mechanism at each step, involving VAE decoding, 3D reconstruction and conditioning injection.
However, the impact on inference speed is minimal, rendering Ouroboros3D efficient for practical applications once training is complete.



%% file: sec/5_con.tex
\section{Conclusion}

In this paper, we introduce Ouroboros3D, a unified framework for single image-to-3D creation that integrates multi-view image generation and 3D reconstruction in a recursive diffusion process. In our framework, these two modules are jointly trained through a self-conditioning mechanism, which allows them to adapt to the inherent characteristic of each stage. By establishing a recursive relationship between these two stages through a 3d aware feedback mechanism, our approach effectively mitigates the data bias encountered in existing two-stage methods.

\section*{Acknowledgment}
This work was supported by National Natural Science Foundation of China (62132001), and the Fundamental Research Funds for the Central Universities.